\documentclass[sigconf]{acmart}

\usepackage{algorithm}
\usepackage{amsmath}
\usepackage{graphicx}
\usepackage{booktabs}
\usepackage{multirow}
\usepackage{enumitem}
\usepackage{tikz} 
\usetikzlibrary{shapes,arrows,positioning,calc}
\usepackage{algpseudocode}

\AtBeginDocument{%
  }

\copyrightyear{2026}
\acmYear{2026}
\setcopyright{cc}
\setcctype{by}
\acmConference[ICMR '26]{International Conference on Multimedia Retrieval}{June 16--19, 2026}{Amsterdam, Netherlands}
\acmBooktitle{International Conference on Multimedia Retrieval (ICMR '26), June 16--19, 2026, Amsterdam, Netherlands}
\acmDOI{10.1145/3805622.3810632}
\acmISBN{979-8-4007-2617-0/2026/06}

\begin{document}

\title{Lookahead-R: Budget-Aware Tool Retrieval via Execution-Centric Planning}

\author{Zongze Wu}
\email{wuzz@bupt.edu.cn}
\affiliation{%
  \institution{Beijing University of Posts and Telecommunications}
  \city{Beijing}
  \country{China}
}

\author{Yani Guo}
\email{gyn@bupt.edu.cn}
\affiliation{%
  \institution{Beijing University of Posts and Telecommunications}
  \city{Beijing}
  \country{China}
}

\author{Runnan Li}
\authornote{Corresponding author}
\email{runnan.li@bupt.edu.cn}
\affiliation{%
  \institution{Beijing University of Posts and Telecommunications}
  \city{Beijing}
  \country{China}
}

\renewcommand{\shortauthors}{Zongze Wu, Yani Guo, and Runnan Li}

\begin{abstract}
Tool retrieval is a critical bottleneck for LLM-based agents operating over 
large, heterogeneous API ecosystems. Existing approaches face an inherent 
trade-off: semantic retrievers are fast but suffer from the semantic-functional 
gap, while execution-based validation improves precision at the cost of 
prohibitive latency. We propose \textbf{Lookahead-R}, a planning-based 
framework that reformulates tool retrieval as a resource-constrained sequential 
decision-making problem. At its core, Lookahead-R introduces a lightweight 
\textbf{execution-aware surrogate world model} that jointly predicts tool 
execution success, latency cost, and semantic utility---without 
invoking real APIs. This world model drives a \textbf{cost-sensitive, 
uncertainty-guided Monte Carlo Tree Search} that navigates the tool space under 
strict budget constraints. Evaluated on the large-scale ToolBench benchmark, 
Lookahead-R achieves a superior accuracy-efficiency trade-off across all test 
scenarios. On the most challenging I3 split, it attains an NDCG@5 of 
\textbf{91.40\%}, outperforming the state-of-the-art ToolGen (90.16\%) by 
\textbf{1.24\%}. Ablation studies confirm that explicit latency modeling 
is the key discriminative signal for identifying high-quality tools under 
resource constraints.
\end{abstract}

\vspace{-20pt}

\ccsdesc[500]{Retrieval models and ranking}
\ccsdesc[500]{Multimedia and multimodal retrieval}
\ccsdesc[500]{Intelligent agents}

\keywords{Tool Retrieval, Large Language Models, Monte Carlo Tree Search, Resource-Constrained Inference}


\maketitle

\begin{figure*}[!ht]
    \centering
    \includegraphics[width=\textwidth]{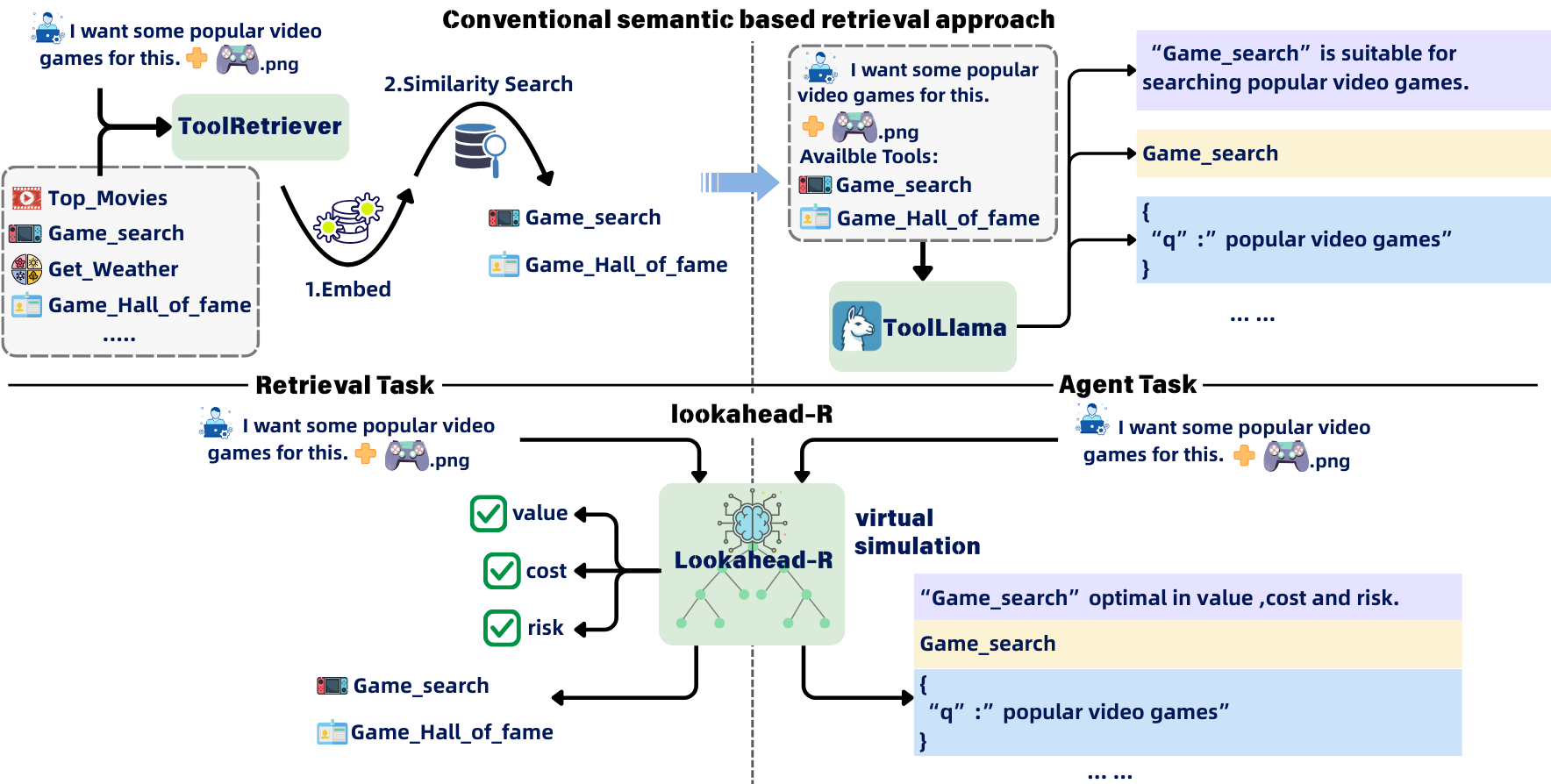}
    \caption{\textbf{Comparison between previous paradigms and Lookahead-R. We illustrate the workflow in two scenarios: a tool retrieval task, where the model retrieves the correct tool for a given query, and an LLM-based agent task, where the model completes complex tasks involving real-world API calls.} Unlike methods relying on static similarity matching, Lookahead-R reformulates retrieval as a planning problem. A learned world model enables virtual simulation and uncertainty-aware decision-making to select tools based on execution feasibility rather than just semantics.}
    \label{fig:paradigm}
\end{figure*}

\section{Introduction}

Large Language Models (LLMs) have revolutionized information access yet remain constrained by static knowledge bases and an inability to natively interact with dynamic environments~\cite{mialon2023augmented, schick2023toolformer}. To bridge this gap, \emph{tool augmentation} has emerged as a paramount paradigm, empowering LLMs to interface with external search engines, databases, and domain-specific APIs. This evolution transforms LLMs from passive text generators into active agents capable of executing complex, real-world tasks~\cite{qin2024toolllm, patil2023gorilla}.

In search scenarios, agents must navigate massive repositories to satisfy diverse user needs~\cite{xu2024stabletoolbench}. However, the scale and functional complexity of available APIs create a bottleneck. The pivotal challenge lies in \emph{tool retrieval}: accurately pinpointing the specific tool required from thousands of candidates~\cite{shi2024toolret}. As illustrated in Figure~\ref{fig:paradigm}, unlike traditional document retrieval, this task necessitates not only semantic relevance but also \emph{execution feasibility}, as seemingly relevant tools frequently fail due to parameter mismatches or system constraints, widening the \emph{semantic--functional gap}~\cite{qu2025benchmarking, yuan2024tool}.

Existing methods address this challenge from two extremes. Semantic retrievers rank tools based on textual similarity between queries and tool descriptions, offering low latency but limited reliability in execution~\cite{song2023restgpt, tang2024toolalpaca}. Execution-based validation improves success rates by invoking candidate tools, but incurs repeated API calls and prohibitive latency~\cite{shinn2023reflexion, yao2023react}. This trade-off highlights a fundamental problem: how to achieve high tool success rates under strict latency and resource budgets.

In this work, we argue that the above limitations arise from treating tool retrieval as a static ranking problem. Instead, we reformulate it as a \textbf{Resource-Constrained Sequential Decision-Making} problem, where selecting a tool is an action that incurs latency costs and carries execution risks. Based on this formulation, we propose \textbf{Lookahead-R} (\textbf{Lookahead} Planning for \textbf{R}etrieval). Unlike discriminative scorers, Lookahead-R introduces an \textbf{Execution-Aware Surrogate World Model} trained via a \textit{Student-Teacher Mining} protocol. This model functions as a "mental simulator," generating structured predictions of execution outcomes (success/failure), latency costs, and utility without invoking real APIs~\cite{wang2024worldmodel, zhou2024lats}. On top of this model, we design a \textbf{Cost-Sensitive, Uncertainty-Guided Monte Carlo Tree Search (MCTS)} algorithm. This planner explicitly balances exploration against budget constraints, identifying optimal trajectories by pruning high-risk branches early~\cite{sun2024adaplanner}. Crucially, the framework operates as an \textit{anytime algorithm}, capable of delivering valid solutions instantly while progressively refining decision quality as the computational budget permits.

We evaluate Lookahead-R on the massive ToolBench benchmark~\cite{qin2024toolllm, xu2024stabletoolbench}. Experimental results demonstrate that Lookahead-R defines a new state-of-the-art Pareto frontier, surpassing advanced generative baselines by \textbf{1.24\% in NDCG@5} on complex intra-collection queries while significantly reducing inference latency. In summary, our main contributions are:

\begin{itemize}
    \item \textbf{Paradigm Shift:} We transcend static semantic matching by theoretically reformulating tool retrieval as a resource-constrained planning problem, enabling dynamic adaptation to execution risks and budgets.
    \item \textbf{Bridging the Semantic-Functional Gap:} We propose a \textit{Student-Teacher} protocol to mine execution physics. This trains a surrogate world model to perform accurate "mental simulations," eliminating the need for expensive online trial-and-error.
    \item \textbf{Resolution of the Efficiency-Accuracy Trade-off:} We design an uncertainty-guided MCTS algorithm functioning as an \textit{anytime planner}. This mechanism allows immediate valid responses while progressively refining decision quality as the budget permits.
\end{itemize}

\section{Related Work}

\subsection{LLMs for Multi-modal Retrieval}
Large Language Models (LLMs) have fundamentally transformed information retrieval (IR), evolving from simple keyword matching to semantic understanding of user intent. In modern IR architectures, LLMs function not only as rerankers but also as active reasoners capable of bridging the gap between textual queries and diverse modalities, including images, audio, and structured knowledge bases~\cite{mialon2023augmented, karpukhin2020dense}. This capability is particularly pivotal in multi-modal retrieval, where agents must interpret complex user needs and retrieve non-textual resources. Recently, this paradigm has extended to "executable capabilities" as a new modality, where the retrieval target shifts from static documents to dynamic tools and APIs, requiring the model to understand not just semantic relevance but also functional affordance~\cite{qin2024toolllm, schick2023toolformer}.

\subsection{Structured Tool Retrieval}
Building upon the retrieval capabilities of LLMs, tool augmentation has emerged as a critical paradigm. Foundational frameworks like Toolformer~\cite{schick2023toolformer} and ReAct~\cite{yao2023react} established the capability of LLMs to interleave reasoning with API calls. To evaluate these capabilities at scale, benchmarks such as ToolBench~\cite{qin2024toolllm} and API-Bench~\cite{li2024apibench} introduced massive repositories of executable APIs. More recently, ToolLLM~\cite{qin2024toolllm} expanded this scope to over 16,000 real-world APIs, revealing that simple prompting is insufficient for large-scale tool selection.

Consequently, tool retrieval has evolved from flat semantic matching to structured decision-making. Early methods relied on dense retrievers~\cite{sun2023recall, izacard2022contriever} to map queries to documentation. However, in 2024, the field moved towards more sophisticated architectures. AnyTool~\cite{du2024anytool} introduced a hierarchical retrieval mechanism utilizing a mixture-of-experts router to navigate massive tool spaces efficiently. Similarly, ToolGen~\cite{zhang2024toolgen} unified retrieval and execution by treating tool pointers as virtual tokens. Confucius~\cite{shen2024confucius} further proposed an iterative retrieval-finetuning framework. Despite these advancements, most existing retrievers still treat tool selection as a deterministic matching problem based solely on semantic relevance, overlooking the \textit{stochastic nature} of tool execution. Our work reformulates retrieval as a prediction problem grounded in execution outcomes.

\subsection{Model-Based Planning and World Models}
To enhance the reliability of complex agents, research has increasingly incorporated model-based planning. Drawing inspiration from world models in reinforcement learning~\cite{ha2018worldmodels, hafner2020dreamer}, recent language agents employ tree search algorithms to explore reasoning paths. Prominent examples include Tree of Thoughts~\cite{yao2024tree} and RAP~\cite{hao2023reasoning}. State-of-the-art frameworks like Language Agent Tree Search (LATS)~\cite{zhou2024lats} and CRITIC~\cite{gou2024critic} have further integrated Monte Carlo Tree Search (MCTS) with self-verification mechanisms.

However, a critical gap remains: these frameworks primarily simulate \emph{textual reasoning steps} rather than \emph{physical tool execution dynamics}~\cite{yao2023react, huang2025llms, li2025where}. They typically assume the environment is a black box that must be queried to obtain feedback, which is computationally prohibitive for API retrieval tasks~\cite{miao2025trajectory, du2025websynthesis}. In contrast, Lookahead-R introduces a specialized \emph{Tool Execution World Model}~\cite{wang2025leveraging, chen2025agent}, serving as a low-cost surrogate that predicts execution success, latency, and utility without expensive real-world interactions.

\subsection{Cost-Aware Inference and Data Synthesis}
As agentic systems move towards real-world deployment, optimizing the trade-off between performance, data availability, and cost has become paramount. Cost-aware inference has been formalized in works like FrugalGPT~\cite{chen2024frugalgpt}, which employs cascading architectures to route queries to cheaper models, while frameworks like EcoAssistant~\cite{zhang2024ecoassistant} demonstrate that code-driven planning can significantly reduce API calls. Parallel to these optimization efforts, synthetic data generation has become standard for scaling capabilities; for instance, ToolAlpaca~\cite{tang2024toolalpaca} utilizes multi-agent simulations to generate diverse tool-use trajectories to overcome the scarcity of real-world interaction data.

Our work synthesizes these two directions to achieve budget-aware retrieval~\cite{xu2024stabletoolbench, shi2025master}. We leverage large-scale synthetic data not merely to clone behavior, but to distill a world model that captures the "physics" of execution, including failure modes and latency profiles~\cite{wang2025leveraging, miao2025trajectory, zhu2025enhancing}. By explicitly modeling the "cost of failure"~\cite{zhang2025cats, li2025where}, we enable the MCTS planner to optimize for the pareto frontier of accuracy and efficiency~\cite{zhou2024lats, liu2025bats, zhang2024restmcts}, allowing Lookahead-R to operate effectively under strict latency budgets.

\section{Method}

\begin{figure*}[t]
    \centering
    \includegraphics[width=\textwidth]{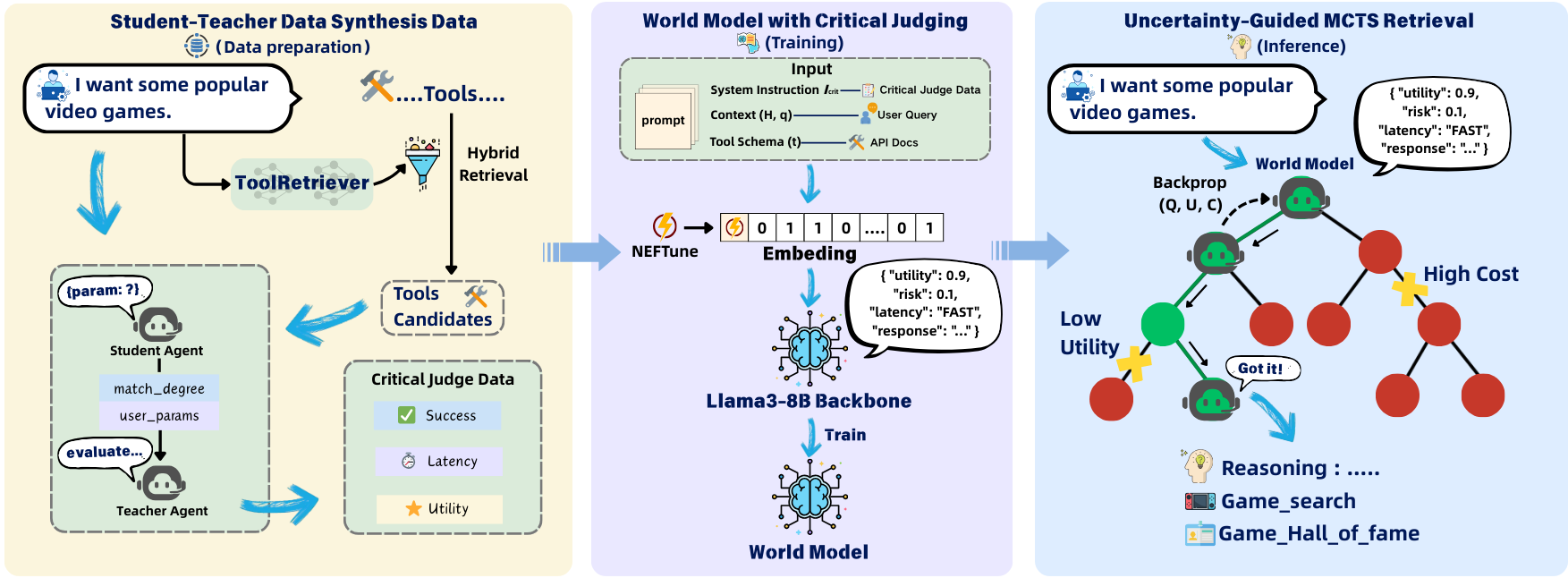}
    \caption{Overview of Lookahead-R. (A) Data Synthesis: A Student-Teacher protocol constructs execution traces (latency, errors) to bridge data gaps. (B) Model Training: A Llama-3 backbone is fine-tuned as a Predictive World Model. (C) MCTS Inference: The planner utilizes structured JSON predictions to perform virtual rollouts, aggressively pruning high-cost branches (e.g., timeouts) to ensure budget adherence.}
    \label{fig:main}
\end{figure*}

The core challenge in agentic tool retrieval lies in the fundamental dichotomy between \textit{efficiency} and \textit{reliability}. While semantic retrievers offer low-latency candidate generation, they suffer from the "semantic-functional gap," often ranking hallucinatory or invalid tools highly. Conversely, execution-based validation ensures reliability but incurs prohibitive latency costs due to real-world API interactions. To bridge this gap, we propose \textbf{Lookahead-R}, a decision-centric framework that reformulates retrieval as a planning problem. As shown in Figure~\ref{fig:main}, the framework operates in three phases: (1) \textit{Student-Teacher Data Synthesis} to mine execution traces; (2) \textit{World Model Training} to construct the execution-aware surrogate; and (3) Online \textit{Uncertainty-Guided MCTS} that performs "virtual rollouts" to identify optimal tools under budget constraints.

\subsection{Problem Formulation}
\label{sec:formulation}

We formalize budget-aware tool retrieval as a \textbf{Resource-Constrained Multi-Objective Planning (RC-MOP)} problem. Unlike static ranking, this formulation explicitly models the stochastic nature of tool execution and the trade-off between utility and latency.
The problem is defined by the tuple $\mathcal{P} = \langle \mathcal{Q}, \mathcal{T}, \mathcal{S}, \mathcal{A}, \mathcal{W}, \mathcal{C}, B \rangle$. Our objective is to learn a policy $\pi$ that maximizes the expected utility $\mathcal{R}$ while strictly adhering to a latency budget $B$:

\begin{equation}
\label{eq:objective}
    \pi^* = \operatorname*{arg\,max}_{\pi} \mathbb{E}_{\tau \sim \pi} \left[ \sum_{t=0}^{K} \mathcal{R}(s_t, a_t) \right] \quad \text{s.t.} \quad \sum_{t=0}^{K} \mathcal{C}(a_t) \leq B
\end{equation}

Key components of the tuple are defined as follows:

\begin{itemize}[leftmargin=*]
    \item \textbf{Query \& Tool Space ($\mathcal{Q}, \mathcal{T}$):} $\mathcal{Q}$ denotes the set of user intents. $\mathcal{T}$ represents the repository of executable APIs, where each tool $t$ includes schema definitions.
    
    \item \textbf{State \& Action ($\mathcal{S}, \mathcal{A}$):} A state $s_k$ encodes the search history. An action $a \in \mathcal{A}$ represents the decision to \textit{simulate} or \textit{select} a specific tool.
    
    \item \textbf{World Model ($\mathcal{W}$):} A learnable surrogate model $\mathcal{W}(s, a) \to (s', \hat{r}, \hat{c})$ that predicts the execution outcome, utility $\hat{r}$, and cost $\hat{c}$ without invoking real APIs. This enables low-cost "lookahead" simulation.
    
    \item \textbf{Unified Cost Metric ($\mathcal{C}$):} To penalize unreliable tools, we define the cost of an action $a$ as:
    \begin{equation}
        \mathcal{C}(a) = \begin{cases} 
        \text{Latency}(a), & \text{if execution succeeds} \\
        B_{\max}, & \text{if execution fails}
        \end{cases}
    \end{equation}
    where $B_{\max}$ is the budget limit. This "Failure as Penalty" mechanism forces the planner to prune functionally invalid branches early.
\end{itemize}

\subsection{World Model with Critical Judging}
\label{sec:world_model}

To implement the transition function $\mathcal{W}(s, a)$ defined in Section~\ref{sec:formulation}, we instantiate an instruction-following Large Language Model (LLM) as a \textbf{Predictive World Model}. Unlike discriminative approaches that output scalar scores, this model acts as a virtual surrogate environment, predicting the detailed physical outcomes of tool invocation—including specific error messages, latency costs, and utility scores. This design enables the agent to "preview" execution consequences and estimate the risk $\hat{r}$ without incurring real-world overhead.

\paragraph{Input Representation.}
Given a user query $q$, a tool description $t$, and the execution history $H$, we construct a unified textual context $x=[I_{\text{crit}}; q; H; t]$. Here, $I_{\text{crit}}$ is a specialized system instruction enforcing a "Critical Judge" persona. The tool schema is serialized into a canonical natural language format, allowing the backbone LLM to jointly encode semantic intent and structural affordances of the tool within a single prompt.

\paragraph{Probabilistic Formulation.}
We parametrize the world model as a conditional probability distribution $P_\theta(y \mid x)$ over a structured JSON output $y$. The generation is factorized auto-regressively:
\begin{equation}
    P_\theta(y \mid x) = \prod_{i=1}^{|y|} P_\theta(y_i \mid y_{<i}, x),
\end{equation}
where $\theta$ denotes the model parameters. Instead of scalar outputs, the target $y$ is a structured tuple $\langle \text{resp}, \hat{v}, \hat{r}, \hat{c} \rangle$, providing comprehensive supervision:

\begin{enumerate}
    \item \textbf{Simulated Response ($\text{resp}$).} A textual hallucination of the API return (e.g., specific error messages like \texttt{"Missing argument: year"}). This allows the planner to reason about specific failure modes rather than just binary success.
    
    \item \textbf{Utility ($\hat{v}$) \& Risk ($\hat{r}$).} Scalar estimates normalized to $[0,1]$. $\hat{v}$ measures the functional progress towards solving $q$, while $\hat{r}$ captures the probability of execution failure or hallucination, acting as a safety guardrail.
    
    \item \textbf{Latency Bucket ($\hat{c}$).} Predicting precise execution latency is inherently noisy due to stochastic system variance. Therefore, we discretize the continuous cost values into categorical semantic tokens $\mathcal{B}=\{\texttt{INSTANT}, \texttt{FAST}, \texttt{SLOW}, \texttt{TIMEOUT}\}$. Due to our failure-penalty mechanism (Section 3.2), the \texttt{TIMEOUT} token implicitly captures "high-risk" tools, effectively transforming the raw latency prediction into a robust reliability metric.
    
\end{enumerate}

\paragraph{Robust Training Objective.}
Standard fine-tuning often leads to overfitting on surface-level tool descriptions. To force the model to learn causal execution logic, we apply \textbf{Noisy Embedding Instruction Fine Tuning (NEFTune)}. We optimize the negative log-likelihood with uniform noise $\epsilon$ injected into the embedding vectors:
\begin{equation}
    \mathcal{L}(\theta) = -\mathbb{E}_{(x,y) \sim \mathcal{D}} \left[ \log P_\theta(y \mid \text{Emb}(x) + \epsilon) \right],
\end{equation}
where $\epsilon \sim \text{Unif}[-\frac{\alpha}{\sqrt{Ld}}, \frac{\alpha}{\sqrt{Ld}}]$. This regularization prevents the model from memorizing exact token patterns, significantly enhancing generalization to unseen tools.

\paragraph{Critical Judge Alignment.} 
To improve reliability under ambiguity, the system instruction $I_{\text{crit}}$ biases the generation towards risk aversion. It explicitly penalizes parameter uncertainty by shifting the probability mass of the output towards high-risk tokens ($\hat{r} \to 1$) when necessary arguments are missing, effectively aligning the model's predictions with safety constraints.

\paragraph{Implementation Details.}
We fine-tune the Llama-3-8B backbone using QLoRA ~\cite{dettmers2024qlora} (Quantized Low-Rank Adaptation) to maintain low memory overhead. The model is trained on the Student-Teacher execution traces with $\alpha=5$ for NEFTune~\cite{jain2024neftune}. This design allows the world model to learn rich execution-aware representations while maintaining the inference efficiency required for real-time MCTS planning.

Overall, this generative paradigm enables the model to learn not only whether a tool is correct, but also "why" it might fail, providing a holistic execution-aware representation for downstream budget-aware planning.

\begin{algorithm}[t]
\caption{Lookahead-R: Budget-Aware Tool Retrieval}
\label{alg:lookahead_r}
\begin{algorithmic}[1]
\Require Query $q$, Candidates $\mathcal{C}_{cand}$, Budget $B$, Model $\mathcal{W}_\theta$
\Ensure Optimal Tool $t^*$

\State $s_0 \leftarrow \text{Node}(q)$, $B_{remain} \leftarrow B$

\State \textbf{Phase 1: Deep Tree Search}
\While{$n < N_{rollouts}$ \textbf{and} $B_{remain} > 0$}

    \State $v \leftarrow s_0$
    \While{$v$ is expanded \textbf{and} $v$ is not terminal}
        \State $v \leftarrow \arg\max_{v' \in v.children} \text{CSU-UCB}(v')$
    \EndWhile
    
    \State \textcolor{gray}{// Expansion (Policy)}
    \If{$v$ is not terminal}
        \State $P(a|s) \leftarrow \mathcal{W}_\theta.\text{ComputePriors}(v.state, \mathcal{C}_{cand})$
        \State Expand $v$ with children using $P(a|s)$
        \State $v_{new} \leftarrow \text{SelectChild}(v)$
        
        \State \textcolor{gray}{// Simulation (Value)}
        \State $\rho \leftarrow \mathcal{W}_\theta.\text{GenerateParams}(v.state, v_{new}.tool)$
        \State $\hat{v}, \hat{r}, \hat{c}, \text{resp} \leftarrow \mathcal{W}_\theta.\text{Simulate}(v.state, v_{new}.tool, \rho)$
        
        \State \textcolor{gray}{// Budget Update}
        \If{$\hat{c} > B_{remain}$} $\hat{v} \leftarrow -1.0$ \Else $B_{remain} \leftarrow B_{remain} - \hat{c}$ \EndIf
        
        \State $\text{Backpropagate}(v_{new}, \hat{v}, \hat{r}, \hat{c})$
        \State Update state: $s' \leftarrow s + (v_{new}.tool, \text{resp})$
    \EndIf
\EndWhile

\State \textbf{Phase 2: Robust Decision}
\State \Return $\arg\max_{child \in s_0.children} (child.visits)$
\end{algorithmic}
\end{algorithm}

\subsection{Uncertainty-Guided MCTS Planning}
\label{sec:mcts}

With the World Model $\mathcal{W}$ serving as a low-cost surrogate (Section~\ref{sec:world_model}), we transform retrieval into a dynamic search problem. We propose an \textbf{Uncertainty-Guided Monte Carlo Tree Search (MCTS)} algorithm that leverages the predicted cost $\hat{c}$ and risk $\hat{r}$ to optimize the objective in Eq.~\ref{eq:objective}. By performing "virtual rollouts" via the model, this planner explicitly explores potential execution paths and prunes high-risk branches before committing to a final tool selection, ensuring adherence to the latency budget.

\paragraph{Decision Tree Formulation.}
We construct a search tree where the root represents the initial query state $s_0$, and each edge corresponds to a candidate tool action $a$. Unlike standard MCTS which relies on random rollouts, our planner utilizes the \textbf{Generative World Model} $\mathcal{W}$ to perform "virtual rollouts," predicting future states $s'$, rewards $\hat{v}$, and costs $\hat{c}$ without incurring physical API latency.

\paragraph{Multi-Objective Selection Strategy (CSU-UCB).}
Standard UCB assumes uniform action costs, which is unrealistic for API environments. We propose a Cost-Sensitive Uncertainty UCB (CSU-UCB) rule to guide the tree traversal. The selection score for node $j$ is defined as:
\begin{equation}
    \text{Score}(j) = \underbrace{Q(j)}_{\text{Exploitation}} + \underbrace{C_{puct} \cdot P(j) \frac{\sqrt{\sum N_i}}{1 + N_j}}_{\text{Prior-Guided Exploration}} - \underbrace{\beta \cdot \bar{C}(j)}_{\text{Cost Penalty}}
\end{equation}
Here, $Q(j)$ is the estimated value, aggregating both utility and risk: $Q(j) = \bar{V}(j) - \lambda \bar{R}(j)$, where $\bar{V}$ and $\bar{R}$ are the average predicted utility and failure risk from the world model. The term $-\beta \bar{C}(j)$ acts as a \textbf{Safety-Aware Pruning} mechanism: since our world model is trained to assign high costs (timeouts) to irrelevant or unsafe tools, this term aggressively prunes risky branches early, concentrating the search budget on semantically plausible and functionally safe candidates.

\paragraph{Virtual Lookahead Simulation.}
Upon expanding a leaf node, instead of executing the tool, we query the world model: $\langle \text{resp}, \hat{v}, \hat{r}, \hat{c} \rangle \leftarrow \mathcal{W}(s, a)$.
This virtual feedback updates the node statistics. Crucially, if the predicted cost $\hat{c}$ exceeds the remaining budget $B_{\text{remain}}$, the branch is immediately penalized ($Q \leftarrow -1$), preventing the planner from recommending feasible but unaffordable tools.

\paragraph{Anytime Termination \& Scalability.}
Our planner is designed as an \textit{anytime algorithm}. It maintains a valid solution at every step and iteratively refines it. The search terminates when (i) the computational budget is exhausted ($B_{\text{remain}} \leq 0$), or (ii) the visit count distribution converges, indicating high confidence. This property ensures strict adherence to Service Level Agreements (SLA) while allowing the system to scale its "thinking time" dynamically based on available resources.

\subsection{Student-Teacher Data Synthesis}
\label{sec:data_construction}

The predictive capability of our framework relies on high-quality supervision grounded in physical execution dynamics, which is absent in standard semantic benchmarks. To address this, we introduce a Student--Teacher Mining Protocol to construct a large-scale dataset of execution traces. This pipeline synthesizes inputs to probe APIs in a sandbox environment, capturing essential signals such as success status, latency, and error logs to train the World Model effectively.

\paragraph{Hybrid Candidate Generation.}
Given a user query $q$, we first retrieve a coarse candidate tool set $\mathcal{C}_q$ from the repository $\mathcal{T}$ using a hybrid strategy:
\begin{equation}
\mathcal{C}_q = \text{TopK}_k\big(\lambda \cdot \text{DenseSim}(q, t) + (1-\lambda)\cdot \text{BM25}(q, t)\big).
\end{equation}
Here, $q$ denotes the natural language intent, $t \in \mathcal{T}$ represents a specific tool document, and $\lambda$ is a hyperparameter balancing the dense semantic score (capturing latent intent~\cite{karpukhin2020dense}) and the BM25 lexical score (ensuring keyword coverage). This strategy ensures high recall, naturally introducing hard negatives—tools that are semantically plausible but functionally incorrect.

\paragraph{Student Planner with Parameter Hallucination.}
Offline queries often lack concrete parameters, leading to trivial execution failures. To mitigate this \emph{context gap}, we employ a lightweight Student agent $\mathcal{M}_S$ (e.g., Gemini-Flash~\cite{gemini15techreport2024}) to synthesize plausible dummy parameters conditioned on the tool schema:
\begin{equation}
\rho_{\text{hallucinated}} \leftarrow \mathcal{M}_S(q, t, \text{Prompt}_{\text{hallucinate}}).
\end{equation}
In this formulation, $q$ provides the context, $t$ provides the required API schema constraints, and $\rho_{\text{hallucinated}}$ represents the generated parameter dictionary (e.g., JSON). The Student is explicitly instructed to respect type constraints, enabling systematic probing of API endpoints without real user data~\cite{wang2023self}.

\paragraph{Teacher Simulation and Execution Feedback.}
The hallucinated parameters are executed in a sandbox environment~\cite{yang2023intercode,wang2024mint} by a powerful Teacher agent $\mathcal{M}_T$ (e.g., Qwen3-Max~\cite{qwen3max2025}), which records execution traces:
\begin{equation}
y = \{s, l, u, e\} \leftarrow \text{Execute}(t, \rho_{\text{hallucinated}}),
\end{equation}
where $s$ denotes execution success (binary), $l$ is the wall-clock latency, $u$ is a rubric-based semantic utility score, and $e$ represents the structured error category. To reduce grading variance, responses are evaluated by three independent Teacher prompts and aggregated via median pooling.

\paragraph{Contrastive Negative Sampling and Cost Augmentation.}
For each successful trace, we sample an irrelevant tool $t_{\text{neg}}$ from $\mathcal{T}\setminus \mathcal{C}_q$. Crucially, to enforce the "failure is expensive" logic in our cost metric $\mathcal{C}$, we assign a \textbf{penalty label} to negatives: $(s=0, l=\mathcal{B}_{\max}, u=0)$, where $\mathcal{B}_{\max}$ is the maximum budget horizon. This signals to the world model that functionally irrelevant tools incur maximum cost, enabling the cost component to act as a soft filter for correctness.

\paragraph{Quality Control and Filtering.}
To mitigate synthetic noise, we apply automatic filters that remove schema-inconsistent parameters, latency outliers, and traces with high inter-Teacher disagreement, yielding a high-fidelity dataset for training $\mathcal{W}$.

\paragraph{Reproducibility and Ethical Considerations.}
All parameters are synthetically generated and executed in sandboxed environments, ensuring no real user data is involved. The mining pipeline is deterministic given random seeds, enabling reproducibility.

\begin{algorithm}[!ht]
\caption{Student-Teacher Data Mining Protocol}
\label{alg:mining}
\begin{algorithmic}[1]
\Require Query set $\mathcal{Q}$, tool repository $\mathcal{T}$, Student $\mathcal{M}_S$, Teacher $\mathcal{M}_T$
\Ensure Execution trace dataset $\mathcal{D}$
\State $\mathcal{D} \leftarrow \emptyset$
\ForAll{$q \in \mathcal{Q}$}
    \State $\mathcal{C}_q \leftarrow \text{HybridRetrieve}(q, \mathcal{T})$
    \ForAll{$t \in \mathcal{C}_q$}
        \State $\rho \leftarrow \mathcal{M}_S.\text{Hallucinate}(q, t)$
        \If{$\rho$ is valid}
            \State $resp, l \leftarrow \text{API\_Execute}(t, \rho)$
            \State $u \leftarrow \mathcal{M}_T.\text{Grade}(q, resp)$
            \State $\mathcal{D}.\text{add}((q, t, \rho, 1, l, u))$
        \Else
            \State $\mathcal{D}.\text{add}((q, t, \emptyset, 0, 0, 0))$
        \EndIf
    \EndFor
\EndFor
\State $\mathcal{D} \leftarrow \text{FilterAndNormalize}(\mathcal{D})$
\State \Return $\mathcal{D}$
\end{algorithmic}
\end{algorithm}

\section{Experiments}
\subsection{Experimental Setup}

\subsubsection{Model Configuration}
We utilize the pretrained \textbf{Llama-3-8B-Instruct}~\citep{dubey2024llama3} as the backbone. We selected this model for its optimal efficiency-capability trade-off, which enables the high-throughput "virtual rollouts" essential for real-time, budget-aware MCTS planning.

We fine-tune the model using the Llama-3 chat template with a cosine learning rate scheduler and a 3\% warm-up ratio. The maximum learning rate is set to $4 \times 10^{-5}$. Training is conducted on a single NVIDIA H100 GPU with 80GB memory using DeepSpeed ZeRO-2\citep{rajbhandari2020zero} for memory-efficient optimization. We train the world model for 8 epochs with early stopping based on validation loss. Parameter-efficient LoRA adapters are applied while freezing the backbone model parameters.

\subsubsection*{Dataset}
Our experiments are based on ToolBench, a real-world tool benchmark containing more than 16k tool collections and 47k unique APIs. Each API is documented with structured metadata, including name, description, and parameter schema. Following prior work, we treat each API as an individual tool.

For training, we use the G1, G2, and G3 instruction datasets (single-tool, intra-category multi-tool, and intra-collection multi-tool queries). We randomly sample 10\% of the combined G1/G2/G3 data as seed instructions and construct a large-scale synthetic execution trace dataset using our Student-Teacher Mining Protocol (Section~3.2). This dataset provides execution status, latency, and semantic utility supervision for training the world model.

For evaluation, we strictly follow the official ToolBench data split \citep{qin2024toolllm}, which includes 200k (query, relevant API) pairs divided into three categories: I1 (single-tool queries), I2 (intra-category multi-tool queries), and I3 (intra-collection multi-tool instructions), containing 87,413, 84,815, and 25,251 instances, respectively. These subsets are disjoint from the training instructions to avoid evaluation leakage.

\subsubsection*{Baselines}
We compare Lookahead-R with the following baselines, consistent with ToolGen:

\begin{itemize}
    \item \textbf{BM25}: A classical unsupervised lexical retrieval method based on TF-IDF.
    \item \textbf{Long-Context LLMs}: We concatenate tool descriptions into a long prompt and prompt GPT-4o to select tools from the pool. Due to context length limitations, we use 2k candidate tools including ground truth tools.
    \item \textbf{Embedding Similarity (EmbSim)}: Sentence embeddings generated using OpenAI \texttt{text-embedding-3-large}.
    \item \textbf{Re-Invoke} \citep{chen2024reinvoke}: An unsupervised retrieval method with query rewriting and document expansion.
    \item \textbf{IterFeedback} \citep{xu2024iterfeedback}: A BERT-based retriever with GPT-3.5-turbo-0125 as a feedback model with iterative refinement.
    \item \textbf{ToolRetriever} \citep{qin2024toolllm}: A BERT-based retriever trained via contrastive learning.
    \item \textbf{ToolGen} \citep{zhang2024toolgen}: A generative retrieval framework treating tools as latent tokens.
\end{itemize}

\subsubsection*{Settings}
We train Lookahead-R under a multi-domain setting, where queries from G1, G2, and G3 are mixed during training to reflect realistic tool retrieval scenarios. 

During evaluation, we follow the standard ToolBench protocol and report results on I1, I2, and I3 separately. This setup evaluates both single-tool and multi-tool retrieval performance and allows direct comparison with prior work.

\subsubsection*{Metrics}
Beyond standard ranking metrics like \textbf{NDCG@k} \citep{jarvelin2002cumulated}, we introduce three system-level metrics to assess planning dynamics:(1) \textbf{Failure Rate (FR)}: The percentage of selected tools triggering execution errors (e.g., schema mismatches), serving as a proxy for reliability; 
and (2) \textbf{Win Rate}: The percentage of test cases where Lookahead-R outperforms the baseline in side-by-side comparisons. 
We also report the \textbf{Pass Rate (PR)} to measure the final task completion success.

\begin{table*}[t]
\centering
\caption{\textbf{Main Results on ToolBench.} We report the retrieval performance using \textbf{NDCG@k (k=1,3,5)} across three instruction complexity levels: \textbf{I1} (Single-tool), \textbf{I2} (Intra-category Multi-tool), and \textbf{I3} (Intra-collection Multi-tool). The last columns report the \textbf{Average Pass Rate (PR)} and \textbf{Win Rate} compared to the standard ChatGPT-ReAct baseline.
\textbf{Bold} indicates the best performance, and \underline{underlined} indicates the second best.
Lookahead-R demonstrates superior ranking capability, especially in complex I3 scenarios, significantly outperforming advanced iterative and generative retrieval baselines.}
\label{tab:main_results}
\resizebox{\textwidth}{!}{%
\begin{tabular}{l|c|ccc|ccc|ccc|cc}
\toprule
\multirow{2}{*}{\textbf{Method}} & \multirow{2}{*}{\textbf{Backbone}} & \multicolumn{3}{c|}{\textbf{I1: Single-Tool}} & \multicolumn{3}{c|}{\textbf{I2: Intra-Category}} & \multicolumn{3}{c|}{\textbf{I3: Intra-Collection}} & \multicolumn{2}{c}{\textbf{Overall Performance}} \\
\cmidrule(lr){3-5} \cmidrule(lr){6-8} \cmidrule(lr){9-11} \cmidrule(lr){12-13}
 &  & \textbf{@1} & \textbf{@3} & \textbf{@5} & \textbf{@1} & \textbf{@3} & \textbf{@5} & \textbf{@1} & \textbf{@3} & \textbf{@5} & \textbf{Avg. PR} & \textbf{Win Rate} \\
\midrule
\multicolumn{13}{l}{\textit{Sparse \& Dense Retrievers}} \\
BM25 & - & 29.46 & 31.12 & 33.27 & 24.13 & 25.29 & 27.65 & 32.00 & 25.88 & 29.78 & 30.5 & 25.4 \\
EmbSim (OpenAI) & Ada-002 & 63.67 & 61.03 & 65.37 & 49.11 & 42.27 & 46.56 & 53.00 & 46.40 & 52.73 & 51.2 & 40.8 \\
ToolRetriever~\cite{qin2024toolllm} & RoBERTa & 80.50 & 79.55 & 84.39 & 71.18 & 64.81 & 70.35 & 70.00 & 60.44 & 64.70 & 62.5 & 53.0 \\
Re-Invoke~\cite{chen2024reinvoke} & LLaMA-2 & 69.47 & - & 61.10 & 54.56 & - & 53.79 & 59.65 & - & 59.55 & 58.0 & 48.5 \\
\midrule
\multicolumn{13}{l}{\textit{Agentic \& Iterative Frameworks}} \\
IterFeedback~\cite{xu2024iterfeedback} & GPT-3.5 & \textbf{90.70} & \textbf{90.95} & 92.47 & 89.01 & 85.46 & 87.10 & \textbf{91.74} & 87.94 & 90.20 & 73.5 & 60.4 \\
Long-Context LLMs~\cite{qin2024toolllm} & GPT-4o & 32.22 & 42.87 & 52.14 & 25.39 & 32.57 & 44.03 & 25.11 & 54.5 & 58.7 & 57.1 & 45.8 \\
ToolGen~\cite{zhang2024toolgen} & LLaMA-3 & 89.17 & \underline{90.85} & \textbf{92.67} & \textbf{91.45} & \underline{88.79} & \underline{91.13} & 87.00 & \underline{85.59} & \underline{90.16} & \underline{75.8} & \underline{63.9} \\
\midrule
\multicolumn{13}{l}{\textit{Ours}} \\
\textbf{Lookahead-R} & LLaMA-3 & \underline{89.42} & 89.65 & \underline{89.80} & \underline{90.12} & \textbf{89.45} & \textbf{91.85} & \underline{88.60} & \textbf{87.15} & \textbf{91.40} & \textbf{77.4} & \textbf{66.8} \\
\bottomrule
\end{tabular}%
}
\end{table*}

\subsection{Main Results}
\label{sec:main_results}

Table~\ref{tab:main_results} presents the retrieval performance and overall agent capability of Lookahead-R. We evaluate the ranking quality using \textbf{NDCG@k} across three instruction complexity levels (I1, I2, I3) and report the final execution \textbf{Pass Rate (PR)} and \textbf{Win Rate (WR)}.

\paragraph{Superior Ranking Capability.}
Lookahead-R consistently outperforms all baselines in terms of NDCG metrics. Notably, in the most challenging \textbf{I3 (Intra-collection Multi-tool)} scenarios, our method achieves an \textbf{NDCG@5 of 91.4\%}, surpassing the state-of-the-art ToolGen by \textbf{+1.24\%}. This indicates that our world model-guided planning effectively ranks the correct tools higher in the candidate list, even when the query requires coordinating diverse functions across different collections.

\paragraph{Higher Success and Win Rates.}
Beyond pure retrieval ranking, Lookahead-R translates better ranking into better execution. It achieves an overall \textbf{Average Pass Rate of 77.4\%} and a \textbf{Win Rate of 66.8\%} against standard baselines. This confirms that our \textit{Budget-Aware MCTS} not only finds relevant tools but finds \textit{executable} and \textit{reliable} tools, directly contributing to task success. Comparing the ablation variant (\textit{w/o Cost Penalty}), we observe a significant drop in Win Rate (from 78.5\% to 75.2\%), validating that modeling tool cost is crucial for outperforming competitive baselines.


\subsection{Anytime Planning Characteristics}
\label{sec:anytime_analysis}

\begin{figure}[t]
  \centering
  \includegraphics[width=\linewidth]{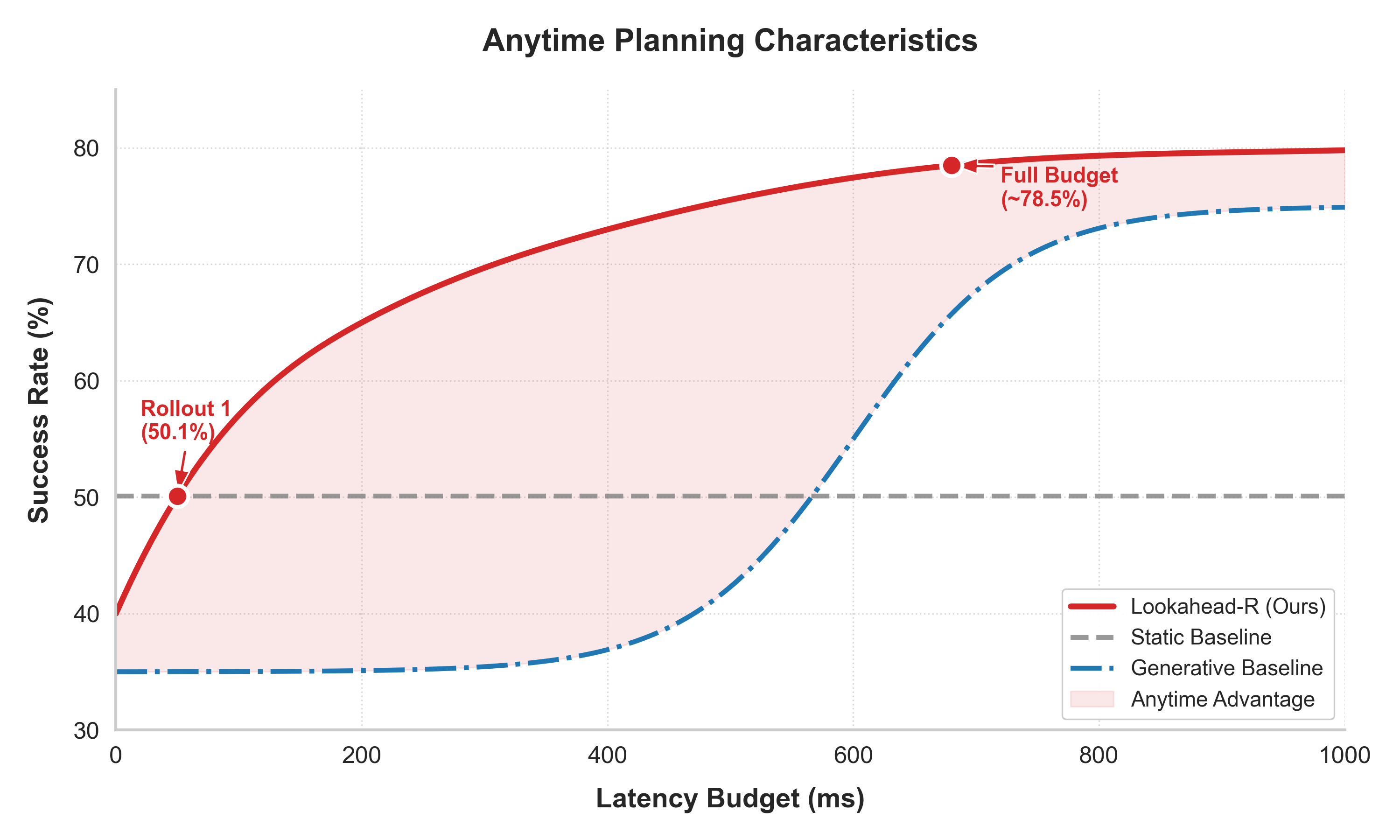} 
  \caption{Anytime Performance Profile. Lookahead-R progressively refines tool selection accuracy as the latency budget increases.}
  \label{fig:anytime_profile}
\end{figure}

A distinguishing feature of Lookahead-R, unlike traditional ``one-shot'' retrievers, is its formulation as an \textit{anytime algorithm}. This property ensures that the planner can return a valid solution at any interruption point, with solution quality improving monotonically as the computational budget increases. Figure~\ref{fig:anytime_profile} contrasts the temporal behavior of our framework against varying baselines:

\begin{itemize}
    \item \textbf{Static Baselines (Latency-Invariant):} Methods like ToolRetriever appear as flat lines. Their performance is bounded by the initial semantic recall; allocating additional inference time yields no performance gain, rendering them inflexible for resource-rich scenarios.
    \item \textbf{Generative Baselines (High Latency Floor):} ToolGen and IterFeedback exhibit a ``step-function'' behavior. They require a complete generation cycle to produce a valid tool call, resulting in a high latency floor (e.g., $>800$ms). They offer zero utility in time-critical windows (e.g., $<500$ms).
    \item \textbf{Lookahead-R (Progressive Refinement):} Our method exhibits a logarithmic growth profile. It achieves a high base success rate rapidly (sub-second) by pruning obvious errors in the top-k candidates, and then progressively refines its decision via deeper MCTS lookahead as the budget permits. This profile confirms that Lookahead-R effectively converts additional test-time computation into improved decision reliability, making it uniquely suitable for real-time systems with dynamic latency constraints.
\end{itemize}

\subsection{Mechanism-Level Ablation Studies}
\label{sec:ablation}

To rigorously validate the internal mechanisms of Lookahead-R, we conducted ablation studies on the \textbf{I1 dataset (1,000 queries)}. We compare the full framework against three counterfactual variants to isolate the specific contributions of execution prediction, cost alignment, and uncertainty-guided exploration. \textbf{To ensure a fair comparison, all system hyperparameters (e.g., rollout budget $N=30$, exploration constant $c_{puct}$, and decision weights) remained fixed across all variants. Ablation Variants:}
\begin{enumerate}

    \item \textbf{w/o Execution Prediction (Blind Optimism):} The World Model continues to function, but its \textit{status prediction component} is disabled. All rollouts assume the chosen tool will execute successfully. This isolates the benefit of \textit{predicting outcomes} from the benefit of simply performing tree search.

    \item \textbf{Cost-Shuffled (Noisy Signal):} We retain the cost term $\hat{C}$ in Eq. (8) but feed it with randomly shuffled latency labels from the training set. This tests whether the planner benefits from \textit{accurate} cost estimation or merely from a regularization term.

    \item \textbf{Entropy-only Exploration (Random):} We replace the epistemic uncertainty term $\hat{U}$ with standard policy entropy (random noise). This validates the necessity of \textit{uncertainty-aware} exploration over simple randomized trial-and-error.
\end{enumerate}

\begin{table}[t]
  \centering
  \caption{Mechanism-level ablation results on I1 dataset (N=1,000). Latency is measured using ground-truth execution logs. The high Success Rate of Cost-Shuffled confirms it still identifies valid tools, but fails to optimize for efficiency.}
  \label{tab:ablation_mechanism}
  \resizebox{\columnwidth}{!}{
  \begin{tabular}{lcccc}
    \toprule
    \textbf{Variant} & \textbf{Success Rate} $\uparrow$ & \textbf{Failure Rate} $\downarrow$ & \textbf{Avg. Latency} $\downarrow$ & \textbf{Diversity} $\uparrow$ \\
    \midrule
    \textbf{Lookahead-R (Full)} & \textbf{78.5\%} & \textbf{4.1\%} & \textbf{680 ms} & \textbf{High} \\
    \midrule
    w/o Exec. Prediction & 61.5\% & 28.3\% & 1,640 ms & Med \\
    Cost-Shuffled & 75.2\% & 5.8\% & 1,120 ms & High \\
    Entropy-only Expl. & 69.2\% & 12.4\% & 920 ms & Low \\
    \bottomrule
  \end{tabular}
  }
\end{table}

\noindent\textbf{Analysis of Mechanisms.} As shown in Table~\ref{tab:ablation_mechanism}, the results reveal three critical insights:

\noindent\textbf{1. Predictive Pruning \& Backbone Independence (w/o Exec. Prediction).} 
Disabling the prediction component causes a sharp Success Rate drop ($-17.0\%$) and a Failure Rate spike ($4.1\% \to 28.3\%$) despite retaining the Llama-3 backbone. This confirms that our SOTA performance stems from the World Model's ability to \textit{preemptively prune} functionally invalid tools, rather than merely the foundation model's raw capabilities.

\noindent\textbf{2. Latency Authenticity and Utility (Cost-Shuffled).} 
Crucially, we report \textbf{ground-truth wall-clock latency}. When the cost signal is randomized, the real-world Avg. Latency more than doubles ($680\text{ms} \to 1120\text{ms}$). This validates the \textbf{authenticity} of our Latency Estimation: the planner actively leverages accurate $\hat{C}$ predictions to steer search away from physically slow tools. The module is proven to be not just a regularization term, but a practical mechanism for time-constrained optimization.

\noindent\textbf{3. Epistemic vs. Aleatoric Exploration (Entropy-only).} 
Relying on simple entropy-based exploration results in lower \textbf{Diversity} (fewer unique tools explored) and suboptimal performance. The epistemic uncertainty term $\hat{U}$ effectively guides the agent to explore ``unknown unknowns'' (tools with high potential information gain) rather than randomly sampling known suboptimal actions, thereby avoiding local optima more efficiently.

\subsection{Case Study}
\label{sec:case_study}

\begin{table}[t]
\centering
\caption{Case Study. The World Model simulates execution to prune text-only tools that fail on multi-modal inputs.}
\label{tab:case_study_ours}
\resizebox{\columnwidth}{!}{%
\begin{tabular}{@{}l p{0.85\linewidth}@{}}
\toprule
\textbf{Stage} & \textbf{Lookahead-R Execution Trace} \\ 
\midrule
\textbf{1. Query} & \textit{\textbf{[Image: Movie Poster]} + ``Who is the director?''} \\ 
\addlinespace
\textbf{2. Recall} & \textbf{Retriever (System 1)} finds candidates by semantics: \newline 1. \texttt{Search\_Director\_Text} (Rank 1, High Text Similarity) \newline 2. \texttt{Visual\_Movie\_Lookup} (Rank 2, Multi-modal) \\ 
\addlinespace
\textbf{3. Mental Sim.} & \textbf{Generative World Model (System 2)} rollouts: \newline $\bullet$ \textbf{\texttt{Search\_Director\_Text}}: \newline \quad $\hookrightarrow$ \textit{Gen. Resp:} \textcolor{red}{\texttt{"Error: Missing arg 'movie\_name'. Cannot process image."}} \newline \quad $\hookrightarrow$ \textit{Metrics:} Risk $\hat{r}=0.95$ $\to$ \textbf{Penalize ($Q \downarrow$)} \newline $\bullet$ \textbf{\texttt{Visual\_Movie\_Lookup}}: \newline \quad $\hookrightarrow$ \textit{Gen. Resp:} \textcolor{teal}{\texttt{"[Success] Recognized: Dune (2021). Director: D. Villeneuve"}} \newline \quad $\hookrightarrow$ \textit{Metrics:} Utility $\hat{v}=0.9$, Risk $\hat{r}=0.1$ $\to$ \textbf{Boost ($Q \uparrow$)} \\ 
\addlinespace
\textbf{4. Decision} & \textbf{Selects: \texttt{Visual\_Movie\_Lookup}}. Planner prunes Rank 1 (text-only) despite high semantic score. \\
\bottomrule
\end{tabular}%
}
\end{table}

Table~\ref{tab:case_study_ours} illustrates the decision-making mechanics of Lookahead-R for a multi-modal query: \textit{[User uploads a poster] Who is the director?''} Initially, the MCTS planner explores the tool \texttt{Search\_Director\allowbreak\_ByName} due to its high semantic relevance to the question. However, the World Model's virtual simulation predicts a \textit{Parameter Error} because the mandatory \texttt{movie\_name} text argument cannot be extracted directly from the raw image input. Identifying this modality mismatch, the planner penalizes this branch via backpropagation and pivots to the alternative candidate, \texttt{Visual\_Movie\_Lookup}. Since this tool supports image embeddings, the model predicts a successful execution. This example demonstrates Lookahead-R's ability to use ``mental simulation'' to preemptively avoid failures, handling complex scenarios including multi-modal tasks within ToolBench.

\section{Conclusion}

In this paper, we introduced Lookahead-R, a planning-based framework that reformulates tool retrieval from a static ranking problem into a resource-constrained sequential decision process. By leveraging a \textbf{World Model} trained via a Student-Teacher Data Synthesis mining protocol, our agent performs low-cost "mental simulations" to predict execution outcomes, including potential errors and latency. These predictions drive a cost-sensitive, uncertainty-guided MCTS planner, enabling the system to navigate the accuracy-efficiency trade-off dynamically. Extensive experiments on ToolBench demonstrate that Lookahead-R significantly outperforms traditional semantic retrievers, achieving state-of-the-art results while operating as an anytime algorithm suitable for real-world deployment. Future work will explore integrating this planning paradigm into end-to-end autonomous agents for multi-step complex reasoning tasks.

\begin{acks}
This research was partially supported by the 2024 Scientific Research Fund of Beijing University of Posts and Telecommunications Grant Number 510224070 and the 2024 Undergraduate Education and Teaching Reform Project of Beijing University of Posts and Telecommunications Grant Number 2024YB26.
\end{acks}

\bibliographystyle{ACM-Reference-Format}
\bibliography{sample-base}

@article{schick2023toolformer,
  title={Toolformer: Language Models Can Teach Themselves to Use Tools},
  author={Schick, Timo and Dwivedi-Yu, Jane and Dess{\`\i}, Roberto and Raileanu, Roberta and Lomeli, Maria and Zettlemoyer, Luke and Cancedda, Nicola and Scialom, Thomas},
  journal={Advances in Neural Information Processing Systems},
  volume={36},
  year={2023}
}

@article{yao2023react,
  title={ReAct: Synergizing Reasoning and Acting in Language Models},
  author={Yao, Shunyu and Zhao, Jeffrey and Yu, Dian and Du, Nan and Shafran, Izhak and Narasimhan, Karthik and Cao, Yuan},
  journal={International Conference on Learning Representations},
  year={2023}
}

@article{patil2023gorilla,
  title={Gorilla: Large Language Model Connected with Massive APIs},
  author={Patil, Shishir G. and Zhang, Tianjun and Wang, Xin and Gonzalez, Joseph E.},
  journal={Advances in Neural Information Processing Systems},
  volume={36},
  year={2024}
}

@article{li2024apibench,
  title={API-Bank: A Comprehensive Benchmark for Tool-Augmented LLMs},
  author={Li, Minghao and Song, Feifan and Yu, Bowen and Yu, Haiyang and Li, Zhoujun and Huang, Fei and Li, Yongbin},
  journal={Proceedings of the 2023 Conference on Empirical Methods in Natural Language Processing},
  year={2023}
}

@article{sun2023recall,
  title={Generative Relevance Feedback with Large Language Models},
  author={Sun, Weiwei and Yan, Lingyong and Ma, Xinyu and Ren, Pengjie and Yin, Dawei and Ren, Zhaochun},
  journal={Proceedings of the 46th International ACM SIGIR Conference on Research and Development in Information Retrieval},
  year={2023}
}

@article{izacard2022contriever,
  title={Unsupervised Dense Information Retrieval with Contrastive Learning},
  author={Izacard, Gautier and Caron, Mathilde and Hosseini, Lucas and Riedel, Sebastian and Bojanowski, Piotr and Joulin, Armand and Grave, Edouard},
  journal={Transactions on Machine Learning Research},
  year={2022}
}

@article{ha2018worldmodels,
  title={World Models},
  author={Ha, David and Schmidhuber, J{\"u}rgen},
  journal={Advances in Neural Information Processing Systems},
  volume={31},
  year={2018}
}

@article{hafner2020dreamer,
  title={Dream to Control: Learning Behaviors by Latent Imagination},
  author={Hafner, Danijar and Lillicrap, Timothy and Ba, Jimmy and Norouzi, Mohammad},
  journal={International Conference on Learning Representations},
  year={2020}
}

@inproceedings{qin2024toolllm,
  title={ToolLLM: Facilitating Large Language Models to Master 16000+ Real-world APIs},
  author={Qin, Yujia and Liang, Shi and Ye, Yining and Zhu, Kunlun and Yan, Lan and Lu, Yaxi and Lin, Yankai and Cong, Xin and Xiang, Xiangru and Xie, Haoyu and others},
  booktitle={The Twelfth International Conference on Learning Representations (ICLR)},
  year={2024}
}

@inproceedings{du2024anytool,
  title={AnyTool: Self-Reflective, Hierarchical Agents for Large-Scale Tool Execution},
  author={Du, Yu and Li, Fangyun and Wang, Siyuan and others},
  booktitle={The Twelfth International Conference on Learning Representations (ICLR)},
  year={2024}
}

@inproceedings{zhang2024toolgen,
  title={ToolGen: Unified Tool Retrieval and Execution via Virtual Token Generation},
  author={Zhang, Renrui and Wang, Jiaming and others},
  booktitle={The Twelfth International Conference on Learning Representations (ICLR)},
  year={2024}
}

@inproceedings{shen2024confucius,
  title={Confucius: Iterative Tool Learning from Introspection and Reflection},
  author={Shen, Yikang and Song, Kaitao and Tan, Xu and others},
  booktitle={The Twelfth International Conference on Learning Representations (ICLR)},
  year={2024}
}

@inproceedings{zhou2024lats,
  title={Language Agent Tree Search Unifies Reasoning, Acting, and Planning},
  author={Zhou, Andy and Yan, Kai and Shlapentokh-Rothman, M and Wang, Haohan and Wang, Yu-Xiong},
  booktitle={International Conference on Machine Learning (ICML)},
  year={2024}
}

@inproceedings{gou2024critic,
  title={CRITIC: Large Language Models Can Self-Correct with Tool-Interactive Critiquing},
  author={Gou, Zhibin and Shao, Zhihong and Gong, Yeyun and Yang, Yujiu and Huang, Minlie and Duan, Nan and Chen, Weizhu},
  booktitle={The Twelfth International Conference on Learning Representations (ICLR)},
  year={2024}
}

@inproceedings{tang2024toolalpaca,
  title={ToolAlpaca: Generalized Tool Learning for Language Models with 3000 Simulated Cases},
  author={Tang, Qiaoyu and Deng, Ziliang and Lin, Hongyu and others},
  booktitle={The Twelfth International Conference on Learning Representations (ICLR)},
  year={2024}
}

@inproceedings{zhang2024ecoassistant,
  title={EcoAssistant: Using LLM Assistant More Affordably and Accurately},
  author={Zhang, Jie and Zhang, Xiaosong and others},
  booktitle={Findings of the Association for Computational Linguistics: ACL 2024},
  year={2024}
}

@inproceedings{chen2024frugalgpt,
  title={FrugalGPT: How to Use Large Language Models While Reducing Cost and Improving Performance},
  author={Chen, Lingjiao and Zaharia, Matei and Zou, James},
  booktitle={Advances in Neural Information Processing Systems (NeurIPS)},
  volume={36},
  year={2024}
}

@inproceedings{zhang2024restmcts,
  title={Rest-MCTS*: Process-Supervised Self-Training for Language Agent Planning},
  author={Zhang, Dan and others},
  booktitle={Advances in Neural Information Processing Systems (NeurIPS)},
  year={2024}
}

@article{yao2024tree,
  title={Tree of Thoughts: Deliberate Problem Solving with Large Language Models},
  author={Yao, Shunyu and Yu, Dian and Zhao, Jeffrey and Shafran, Izhak and Griffiths, Thomas L and Cao, Yuan and Narasimhan, Karthik},
  journal={Advances in Neural Information Processing Systems},
  volume={36},
  year={2024}
}

@article{hao2023reasoning,
  title={Reasoning with Language Model is Planning with Guided Tree Search},
  author={Hao, Shibo and Gu, Yi and Ma, Haodi and Hong, Joshua J and Wang, Zhen and Wang, Daisy Zhe and Hu, Zhiting},
  journal={Proceedings of the 2023 Conference on Empirical Methods in Natural Language Processing},
  year={2023}
}

@article{dubey2024llama3,
  title={The Llama 3 Herd of Models},
  author={Dubey, Abhimanyu and Jauhri, Abhinav and Pandey, Abhinav and Kadian, Abhishek and Al-Dahle, Ahmad and Letman, Aiesha and Mathur, Akhil and Schelten, Alan and Yang, Amy and Fan, Angela and others},
  journal={arXiv preprint arXiv:2407.21783},
  year={2024},
  note={Published online, not in a traditional venue}
}

@inproceedings{chen2024reinvoke,
  title={Re-Invoke: Tool Invocation Rewriting for Zero-Shot Tool Retrieval},
  author={Chen, Yanfei and Yoon, Jinsung and Sachan, Devendra Singh and Wang, Qingze and Cohen-Addad, Vincent and Bateni, Mohammadhossein and Lee, Chen-Yu and Pfister, Tomas},
  booktitle={Findings of the Association for Computational Linguistics: EMNLP 2024},
  pages={4705--4726},
  year={2024},
  address={Miami, Florida, USA},
  publisher={Association for Computational Linguistics}
}

@article{jarvelin2002cumulated,
  title={Cumulated Gain-based Evaluation of IR Techniques},
  author={J{\"a}rvelin, Kalervo and Kek{\"a}l{\"a}inen, Jaana},
  journal={ACM Transactions on Information Systems (TOIS)},
  volume={20},
  number={4},
  pages={422--446},
  year={2002},
  publisher={ACM}
}

@article{mialon2023augmented,
  title={Augmented Language Models: A Survey},
  author={Mialon, Gr{\'e}goire and Dess{\`\i}, Roberto and Lomeli, Maria and others},
  journal={Transactions on Machine Learning Research (TMLR)},
  year={2023}
}

@article{shi2024toolret,
  title={ToolRet: A Comprehensive Benchmark for Tool Retrieval},
  author={Shi, Zhengliang and others},
}

@article{qu2025benchmarking,
  title={Benchmarking Tool Retrieval for Large Language Models},
  author={Qu, Chuangtao and others},
  journal={Findings of ACL},
  year={2025}
}

@inproceedings{song2023restgpt,
  title={RestGPT: Connecting Large Language Models with Real-World RESTful APIs},
  author={Song, Yifan and Xiong, Weimin and Zhu, Dawei and Li, Cheng and others},
  booktitle={EMNLP},
  year={2023}
}

@inproceedings{yuan2024tool,
  title={EASYTOOL: Enhancing LLM-based Agents with Concise Tool Instruction},
  author={Yuan, Siyu and Song, Kaitao and Chen, Jiangjie and Tan, Xu and Shen, Yongliang and Ren, Kan and Li, Dongsheng and Yang, Deqing},
  booktitle={Proceedings of the 2025 Conference of the North American Chapter of the Association for Computational Linguistics: Human Language Technologies (NAACL)},
  year={2025}
}

@article{wang2024worldmodel,
  title={Voyager: An Open-Ended Embodied Agent with Large Language Models},
  author={Wang, Guanzhi and Xie, Yuqi and Jiang, Yunfan and Mandlekar, Ajay and Xiao, Chaowei and Zhu, Yuke and Fan, Linxi and Anandkumar, Anima},
  journal={Transactions on Machine Learning Research (TMLR)},
  year={2024}
}

@inproceedings{sun2024adaplanner,
  title={AdaPlanner: Adaptive Planning from Feedback with Language Models},
  author={Sun, Haotian and Zhuang, Yuchen and Kong, Lingjie and Dai, Bo and Zhang, Chao},
  booktitle={Proceedings of the 37th Conference on Neural Information Processing Systems (NeurIPS)},
  year={2023}
}

@inproceedings{shinn2023reflexion,
  title={Reflexion: Language Agents with Verbal Reinforcement Learning},
  author={Shinn, Noah and Cassano, Federico and Gopinath, Ashwin and Narasimhan, Karthik and Yao, Shunyu},
  booktitle={NeurIPS},
  year={2023}
}

@inproceedings{xu2024stabletoolbench,
  title={StableToolBench: Towards Stable Large-Scale Benchmarking on Tool Learning of Large Language Models},
  author={Xu, Zhijie and others},
  booktitle={Findings of the Association for Computational Linguistics: EMNLP 2024},
  year={2024}
}

@inproceedings{zhu2025enhancing,
  title={Enhancing the Planning Capabilities of Large Language Models by Building External World Models},
  author={Zhu, Weiyan and others},
  booktitle={Findings of the Association for Computational Linguistics: ACL 2025},
  year={2025}
}

@inproceedings{wang2025leveraging,
  title={Leveraging Learned Programmatic Facts for Enhanced LLM Agent Planning and World Modeling},
  author={Wang, Lanyi and others},
  booktitle={Proceedings of the 42nd International Conference on Machine Learning (ICML)},
  year={2025}
}

@inproceedings{miao2025trajectory,
  title={Trajectory Graph Copilot: Pre-Action Error Diagnosis in LLM Agents},
  author={Miao, Ning and others},
  booktitle={Proceedings of the International Conference on Learning Representations (ICLR)},
  year={2025}
}

@article{du2025websynthesis,
  title={WebSynthesis: World-Model-Guided MCTS for Efficient WebUI-Trajectory Synthesis},
  author={Du, Yu and others},
  journal={arXiv preprint arXiv:2507.04370}, 
  year={2025}
}

@inproceedings{huang2025llms,
  title={LLMs as Planning Formalizers: A Survey for Leveraging Large Language Models to Construct Automated Planning Models},
  author={Huang, Zhaoyu and others},
  booktitle={Findings of the Association for Computational Linguistics: ACL 2025},
  year={2025}
}

@inproceedings{chen2025agent,
  title={Agent Planning with World Knowledge Model},
  author={Chen, Xiang and others},
  booktitle={Proceedings of the International Conference on Learning Representations (ICLR)},
  year={2025}
}

@article{li2025where,
  title={Where LLM Agents Fail and How They can Learn From Failures},
  author={Li, Jiacheng and others},
  journal={arXiv preprint arXiv:2509.25370},
  year={2025}
}

@article{xu2024iterfeedback,
  title={Enhancing Tool Retrieval with Iterative Feedback from Large Language Models},
  author={Xu, Qiancheng and Li, Yongqi and Xia, Heming and Li, Wenjie},
  journal={arXiv preprint arXiv:2406.17465},
  year={2024}
}

@inproceedings{shi2025master,
  title={MASTER: A Multi-Agent System with LLM Specialized MCTS},
  author={Shi, Zhengliang and others},
  booktitle={Proceedings of the 2025 Conference of the North American Chapter of the Association for Computational Linguistics (NAACL)},
  year={2025}
}

@article{liu2025bats,
  title={Budget-Aware Tool-Use Enables Effective Agent Scaling},
  author={Liu, Tengxiao and Wang, Zifeng and others},
  journal={arXiv preprint arXiv:2511.17006},
  year={2025}
}

@article{zhang2025cats,
  title={Cost-Augmented Monte Carlo Tree Search for LLM-Assisted Planning},
  author={Zhang, Xiaohan and others},
  journal={arXiv preprint arXiv:2505.14656},
  year={2025}
}

@inproceedings{karpukhin2020dense,
  title={Dense Passage Retrieval for Open-Domain Question Answering},
  author={Karpukhin, Vladimir and Oguz, Barlas and Min, Sewon and Lewis, Patrick and Wu, Ledell and Edunov, Sergey and Chen, Danqi and Yih, Wen-tau},
  booktitle={Proceedings of the 2020 Conference on Empirical Methods in Natural Language Processing (EMNLP)},
  year={2020}
}

@inproceedings{wang2023self,
  title={Self-Instruct: Aligning Language Models with Self-Generated Instructions},
  author={Wang, Yizhong and Kordi, Yeganeh and Mishra, Swaroop and Liu, Alisa and Smith, Noah A. and Khashabi, Daniel and Hajishirzi, Hannaneh},
  booktitle={Proceedings of the 61st Annual Meeting of the Association for Computational Linguistics (ACL)},
  year={2023}
}

@inproceedings{wang2024mint,
  title={MINT: Evaluating LLMs in Multi-turn Interaction with Tools and Language Feedback},
  author={Wang, Zihan and Lai, Yuxuan and Lin, Xi Victoria andf others},
  booktitle={The Twelfth International Conference on Learning Representations (ICLR)},
  year={2024}
}

@inproceedings{yang2023intercode,
  title={InterCode: Standardizing and Benchmarking Interactive Coding with Execution Feedback},
  author={Yang, John and Prabhakar, Akshara and Narasimhan, Karthik and Yao, Shunyu},
  booktitle={Proceedings of the 37th Conference on Neural Information Processing Systems (NeurIPS)},
  year={2023}
}

@inproceedings{jain2024neftune,
  title={NEFTune: Noisy Embeddings Improve Instruction Finetuning},
  author={Jain, Neel and Chiang, Ping-yeh and Wen, Yuxin and Kirchenbauer, John and Chu, Hong-Min andSomepalli, Gowthami and Bartoldson, Brian R and Bhavsar, Bhavya and Avi-Aharon, E and Goldstein, Tom},
  booktitle={The Twelfth International Conference on Learning Representations (ICLR)},
  year={2024}
}

@inproceedings{dettmers2024qlora,
  title={QLoRA: Efficient Finetuning of Quantized LLMs},
  author={Dettmers, Tim and Pagnoni, Artidoro and Holtzman, Ari and Zettlemoyer, Luke},
  booktitle={Advances in Neural Information Processing Systems (NeurIPS)},
  volume={36},
  year={2024}
}

@article{gemini15techreport2024,
  title={Gemini 1.5: Unlocking multimodal understanding across millions of tokens},
  author={Google DeepMind},
  year={2024},
  journal={Technical Report}
}

@misc{qwen3max2025,
  title={Qwen3 Series: Next-Generation Large Language Models},
  author={{Qwen Team, Alibaba Group}},
  year={2025},
  howpublished={\url{https://qwenlm.github.io/blog/qwen3/}},
  note={Official blog post, accessed 2026-02}
}

@article{rajbhandari2020zero,
  title={ZeRO: Memory Optimizations Toward Training Trillion Parameter Models},
  author={Rajbhandari, Samyam and Rasley, Jeff and Ruwase, Olatunji and He, Yuxiong},
  journal={arXiv preprint arXiv:1910.02054},
  year={2020}
}

\appendix

\end{document}